\documentclass{article}
\usepackage{ijcai24}

\usepackage{times}
\usepackage{soul}
\usepackage{url}
\usepackage[hidelinks]{hyperref}
\usepackage[utf8]{inputenc}
\usepackage[small]{caption}
\usepackage{graphicx}
\usepackage{amsmath}
\usepackage{amsfonts}
\usepackage{amsthm}
\usepackage{booktabs}
\usepackage{algorithm}
\usepackage{algorithmic}
\usepackage[switch]{lineno}
\usepackage{bm}
\usepackage{eucal}
\title{Generating Part-Based Global Explanations Via Correspondence}

\author{
Kunal Rathore
\and
Prasad Tadepalli
\affiliations
Oregon State University\\
\emails
\{rathorek, tadepall\}@oregonstate.edu \\
}
\begin{document}

\maketitle

% -------------------------------------------------------------------
% ===================================================================

\begin{abstract}
    Deep learning models are notoriously opaque. Existing explanation methods often focus on localized visual explanations for individual images. Concept-based explanations, while offering global insights, require extensive annotations, incurring significant labeling cost. We propose an approach that leverages user-defined part labels from a limited set of images and efficiently transfers them to a larger dataset. This enables the generation of global symbolic explanations by aggregating part-based local explanations, ultimately providing human-understandable explanations for model decisions on a large scale. 
\end{abstract}

% -------------------------------------------------------------------
% ===================================================================
\section{Introduction}

Deep Neural Networks (DNNs) have achieved impressive results in various fields, including medical diagnosis \cite{bakator2018deep}, computational medicine \cite{yang2021intelligent}, self-driving cars \cite{app10082749}, and safety-critical industries \cite{8897630}. However, their growing complexity raises concerns about their ``black-box" nature, hindering user understanding and trust. This necessitates the development of explainable methods for DNNs, particularly in human-interactive and safety-critical contexts.

Saliency maps are prominent tools for producing post hoc explanations for Convolution Neural Network (CNN) image classification models, often generated using gradient or perturbation-based approaches \cite{springenberg,simonyan,zeiler,selvaraju}. The gradient-based approach leverages the gradients of the model's output with respect to the input image, and the magnitude of gradients is used to indicate the pixel-level importance in an image. Shitole et al.~\shortcite{Shitole} argue that a single saliency map provides an incomplete understanding, since there are often many other maps that can explain a classification equally well. They employ a beam search algorithm to systematically search for multiple \textit{minimal sufficient explanations} (MSXs) for each image organized into a structured attention graph (SAG). However, this explanation approach faces two major limitations; a) MSXs only provide explanations for a single image and need other approaches for global insights; b) SAGs offer limited user-friendliness, requiring more time for comprehension.

% Collapse multiple citations as
% follows:~\cite{gls:hypertrees,levesque:functional-foundations}.
% \nocite{abelson-et-al:scheme}
% \nocite{bgf:Lixto}
% \nocite{brachman-schmolze:kl-one}
% \nocite{gottlob:nonmon}
% \nocite{gls:hypertrees}
% \nocite{levesque:functional-foundations}

% ==============================================================
\begin{figure}[t]
  \centering
    \includegraphics[width=0.9\columnwidth]{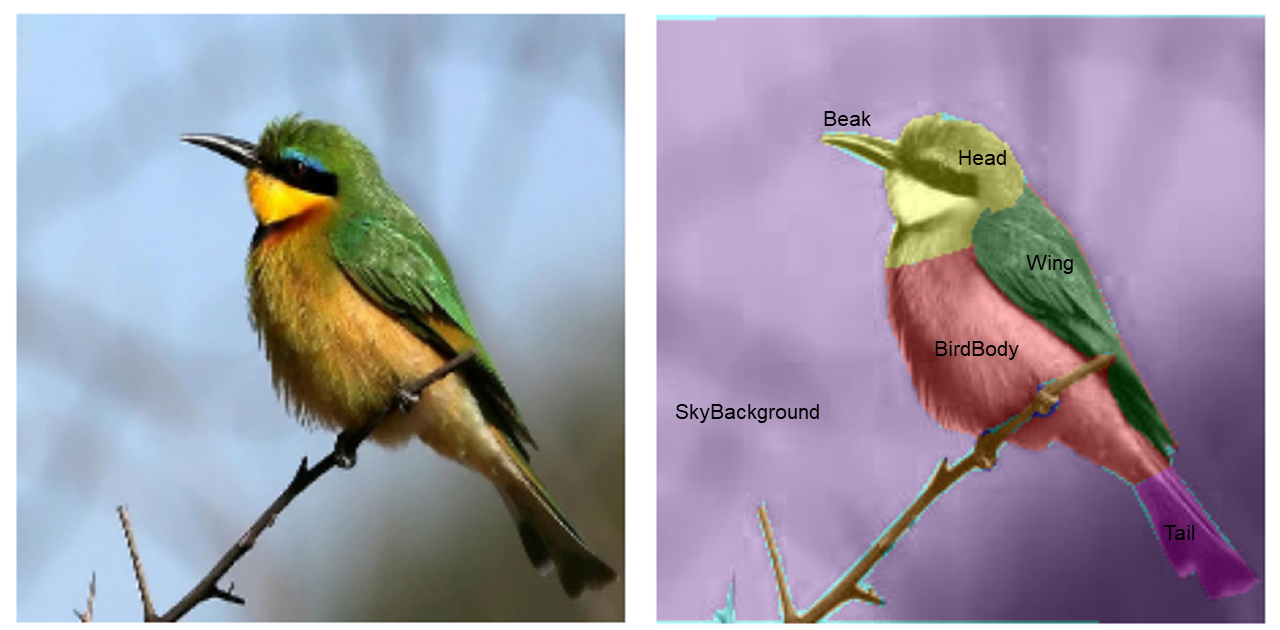}
   \caption{Image of a Bee-Eater from PartImageNet dataset (left) and annotation of the same image with body part labels (right).}
   \label{fig:parts}
\end{figure}
% --------------------------------------------------------------
Based on the scope of interpretability, explanation methods are categorized into local \cite{baehrens2010explain,ribeiro2016should} and global explanations \cite{pmlr-v80-kim18d,lakkaraju2016interpretable}. \textbf{Local} explanations expose the parts of the input responsible for the model predictions on a specific instance. %These explanations are handy when we are interested in understanding a specific instance. 
In contrast, \textbf{global} explanations focus on identifying general decision patterns that apply to an entire dataset.  For instance, global explanations can indicate to what extent the long-heavy curved beak is responsible in identifying  a Wood Stork bird category by a given model. The global explanations can aid in evaluating, validating, and building trust in AI models. 
In this paper, we address the problem of generating succinct post hoc global explanations for the predictions of the model in terms of the object parts. A brute-force solution involves manually 
mapping the local image segment-level explanations into object parts, which is labor-intensive. 
Our approach takes advantage of the fact that the deep features learned by the model contain high-level concepts such as parts of the object \cite{erhan2009visualizing}. Hence, we employ a multilayer CNN feature matching approach called the Hyperpixel Flow (HPF) ~\cite{Min_2019_ICCV} that matches the part labels of a small number of annotated images with unlabeled query images. However, the HPF algorithm works well only on visually similar images. Hence, before invoking it, we extract visually similar images for the query image from the annotated set of images.

In the next phase, we create local explanations using a beam search algorithm \cite{Shitole}. These explanations are subsequently transformed into part-based expressions using the part matching found in the previous step. Finally, global explanations on the dataset are derived by employing the Greedy Set Cover algorithm \cite{young2008greedy}. In summary, our contributions are as follows.

\begin{itemize}
\item We implemented a system called GEPC (Global Explanations via Part Correspondence) to produce part-based global explanations of an image classification model by combining local explanation search with part correspondence and greedy set cover.
\item We employed a novel training-set, test-set methodology for evaluating global explanations, and used it to show the effectiveness of our approach on multiple datasets.  
\end{itemize}

% ===================================================================

\section{Related Work}
\label{sec:related-work}

Explanation methods in the literature are sometimes characterized by the chronology of the explanation step w.r.t. the model training. {\em Inherent} or {\em intrinsic} explanation methods provide explanations while training, while {\em post hoc} methods explain any pre-trained neural network's decisions \cite{camburu2019can}.

Some explanation methods are {\em gradient-based} in that they 
compute gradients of the output score of different units w.r.t. the input image \cite{selvaraju}, while others are {\em perturbation-based}, i.e., they alter parts of the input to determine the effect of change on the model decision, subsequently representing the contribution of each part \cite{fong2017interpretable}. Other methods such as LIME \cite{zhang2019should} and ANCHORS \cite{ribeiro2018anchors} build local surrogate models in lieu of explanations. 

{\em Concept-based} explanation methods aim to align user-defined concepts with the model’s decision.
%These concepts can be pre-annotated data or learned from another model. 
Techniques such as TCAV \cite{pmlr-v80-kim18d}, Net2Vec \cite{fong2018net2vec}, and IBD \cite{zhou2018interpretable} interpret the internal states of deep neural networks in terms of high-level concepts.

% --------------------------------------------------------------------------------

Recently, \cite{NEURIPS2022_ddb8486b} proposed that visual correspondence-based explanations improve their robustness. This motivates us to explore the potential of using correspondence to generate part-based concepts on larger datasets. 
Part-based object recognition is a classical approach in which the objective is to collect information at the local level and to form a classification model \cite{yan2015object,shu2013improving,felzenszwalb2009object}. Despite many implementations for tasks such as segmentation, classification, and object detection, part-based approaches have never been explored for generating post hoc human-readable explanations.

% -------------------------------------------------------------------
% ===================================================================
 \section{Global Explanations via Part Correspondence}
\label{sec:approach}

\begin{figure}[t]
\centering
% width=0.9\columnwidth]
\includegraphics[width=0.9\columnwidth]{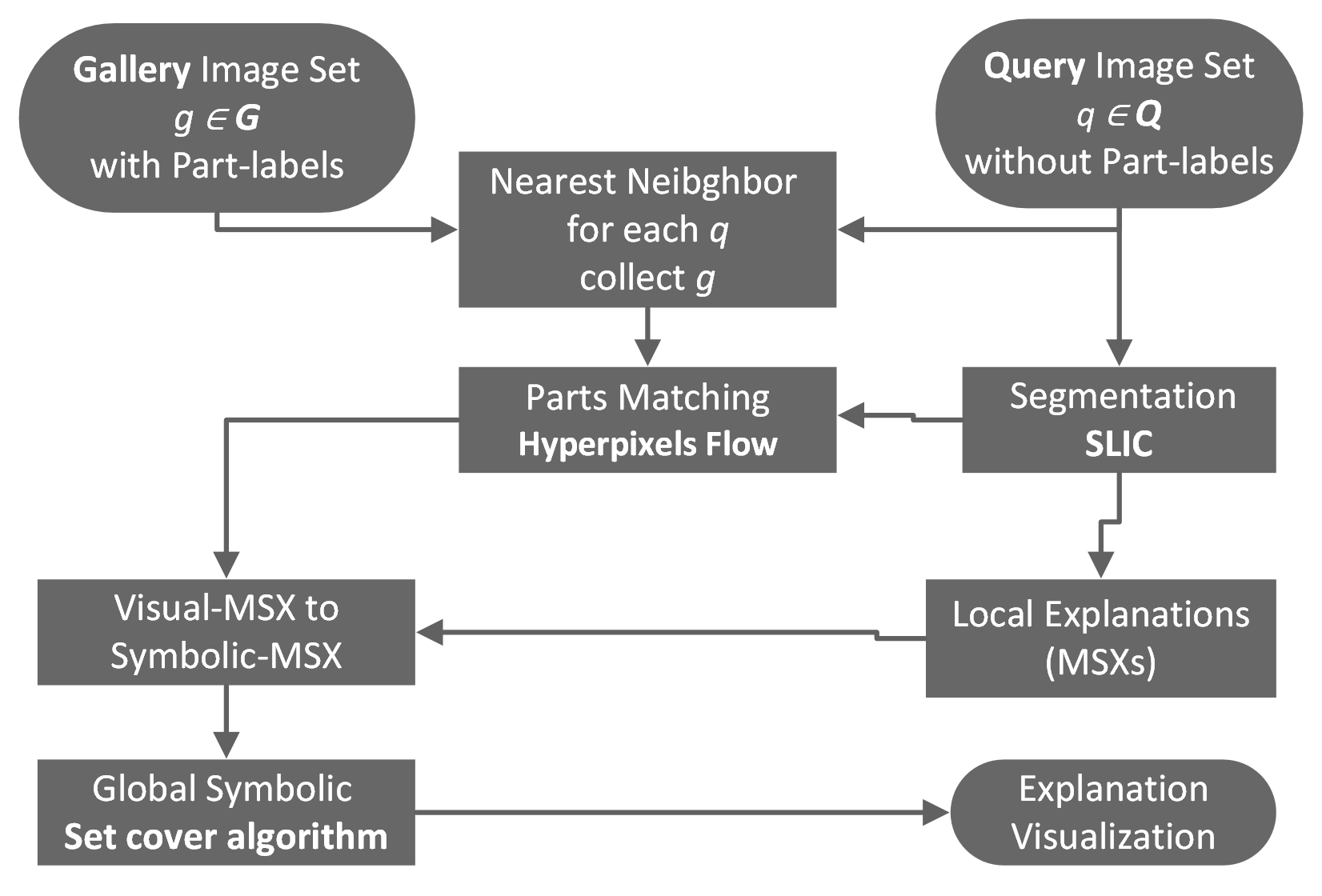} 
\caption{Overall flow diagram for the proposed method. This consists of five steps: unsupervised segmentation, computing nearest neighbor, part correspondence, computing local explanations, and extracting global symbolic explanations.}

\label{fig:framework}
\end{figure}

Our GEPC approach consists of deriving multiple local explanations for each image, mapping them to object parts, and finding a succinct global explanation that covers the predictions of the model on the entire dataset. 
%as a decision list of part-based expressions. 
The system pipeline is illustrated in Figure~\ref{fig:framework} and consists of image segmentation, finding a visually similar annotated gallery image for each query image, corresponding parts of the query image to the gallery image, finding local explanations for each image by beam search and deriving a global explanation from the local explanations via greedy search. These steps are detailed in the rest of this section.  

% ============================================================

% ==============================================================
\begin{figure}[t]
  \centering
   %\includegraphics[width=0.8\linewidth]{egfigure.eps}
    \includegraphics[width=0.9\columnwidth]{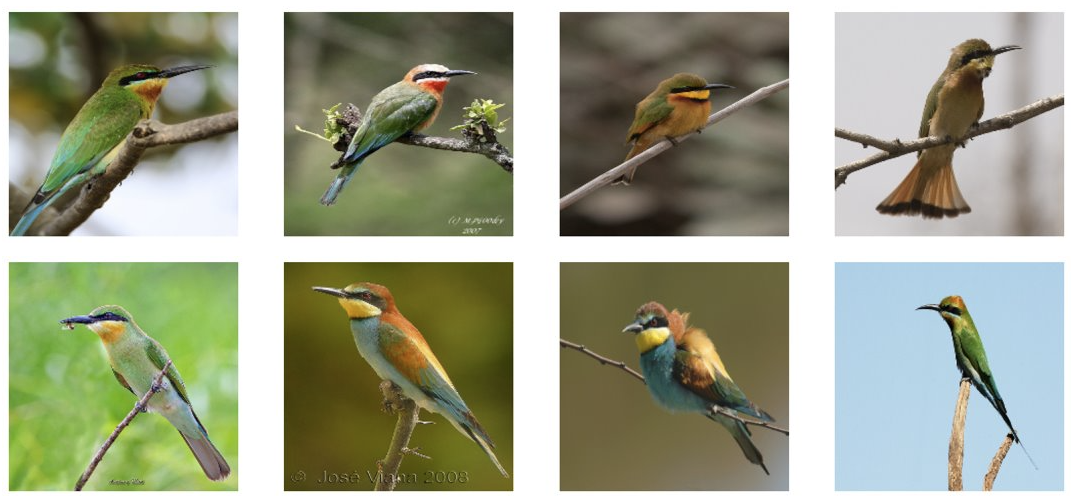}
   \caption{Visual similarity (across the row), these images belong to the Bee-Eater category from the ImageNet dataset. Each row shows images having same viewing angle and same object orientations.}
   \label{fig:visual_sim}
\end{figure}

% ==============================================================
\begin{figure}
  \centering
   %\includegraphics[width=0.8\linewidth]{egfigure.eps}
    \includegraphics[width=0.9\columnwidth]{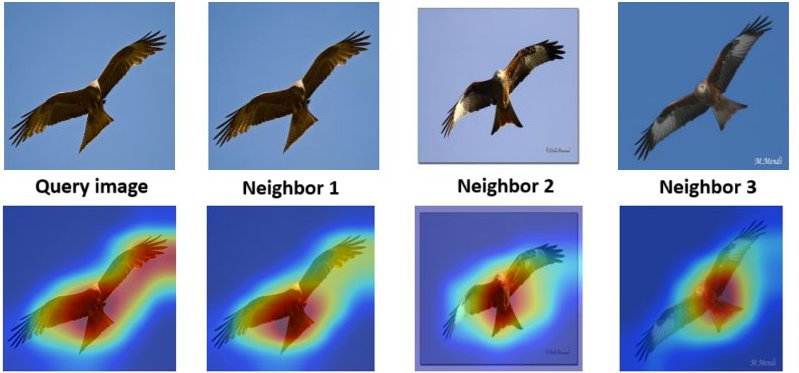}
   \caption{Row 1: the closest neighbors to the query image in the 
   Kite category from ImageNet dataset; Row 2: the upsampled output as the summed weights $\sum_{l} {w}^l$  overlapped on the corresponding images.}
   \label{fig:heatmaps}
\end{figure}
%-------------------------------------------------------------------------

\begin{figure}
  \centering
    \includegraphics[width=0.9\columnwidth]{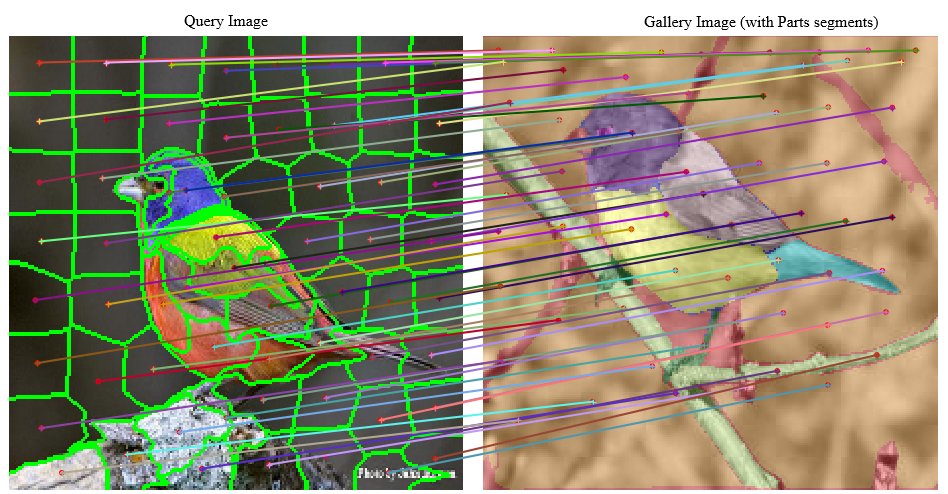}
   \caption{Correspondence: Matching spatial features over images, here centroid of each superpixel from query image are queried for matching pixel on gallery image (parts annotated) }
   \label{fig:correspond}
\end{figure}

%-------------------------------------------------------------------------

\subsection{Part correspondence}

There exist several methods for label transfer, typically done by matching features in images and known as the feature / keypoint correspondence \cite{ma2021image}. 

Due to performance reasons, most of these methods require supervised or weakly supervised learning. However, since we focus on a post hoc approach using a pre-trained model, we employ the Hyperpixel flow \cite{Min_2019_ICCV} technique that helps map the learned high-level features across visually similar images.

HPF performs matching over the deep features, motivated by the fact that different layers of the DNN model learn low-level features to high-level concepts in images. This method is inspired by the hypercolumns \cite{hariharan2015hypercolumns} used in object segmentation and detection. They illustrated that hypercolumns that combine features from multiple layers of CNN improve object detection and part labeling.

Given an image, CNN's intermediate outputs $(\mathbf{f}^0, \mathbf{f}^1,…, \mathbf{f}^{L-1})$ are pooled and upsampled to create a hyperimage, combining $Z$ spatial features.
\begin{equation}
    \mathbf{F} = [\mathbf{f}^{l_0}, \mathbf{U}(\mathbf{f}^{l_1}), \mathbf{U}(\mathbf{f}^{l_2}) ..., \mathbf{U}(\mathbf{f}^{l_{Z-1}})]
\end{equation}
where $\mathbf{U}$ denotes a function that upsamples the input feature map to the size of $\mathbf{f}^{l_0}$, the base map.
Each spatial position $p$ in the hyperimage corresponds to image coordinates $\mathbf{x}_p$ and a hyperpixel feature $\mathbf{f}_p$, where $\mathbf{f_p = F}(x_p)$. The hyperpixel at position $p$ is denoted as $\mathbf{h}_p = (\mathbf{x}_p, \mathbf{f}_p)$.  The key idea of matching is to re-weight appearance similarity by Hough space voting to enforce geometric consistency. Let $\mathcal{D} = (\mathcal{H}, \mathcal{H}')$ be two sets of hyperpixels, and $m = (h, h')$ be a hyperpixel match where $h$ and $h'$ are respectively elements of $\mathcal{H}$ and $\mathcal{H}'$. Given a Hough space $\chi$ of possible offsets between the two hyperpixels, the confidence for match $m$, is computed as: 
\begin{equation}
    p(m|D) \propto p(m_a) \sum_{x \in \chi} p(m_g|\mathbf{x})  \sum_{m \in \mathcal{H} \times \mathcal{H}' } p(m_a) p(m_g|\mathbf{x})
\end{equation}

where $p(m_a)$ represents the confidence in the appearance matching and $p(m_g|\mathbf{x})$ is the confidence for geometric matching with an offset $\mathbf{x}$, measuring how close the offset induced by $m$ is to $\mathbf{x}$. Here, appearance matching confidence is derived using exponentiated cosine distance over hyperpixel features.
\begin{equation}
    p(m_a)  = \textrm{ReLU} \left(\frac{\mathbf{f.f'}}{\mathbf{||f|| ||f'||}} \right)^d
\end{equation}

We employ the HPF algorithm \cite{Min_2019_ICCV}, which leverages the features learned from the model in multiple layers to match the corresponding parts in the images. SLIC \cite{achanta2012slic} technique is used to generate segments of each query image. Centroids $(C_x, C_y)$ of each query image segment act as predictor keypoints in Hyper-pixel flow, enabling matching corresponding regions to transfer parts from the gallery image to the query image.

%-------------------------------------------------------------------

\begin{figure}
  \centering
    \includegraphics[width=0.9\columnwidth]{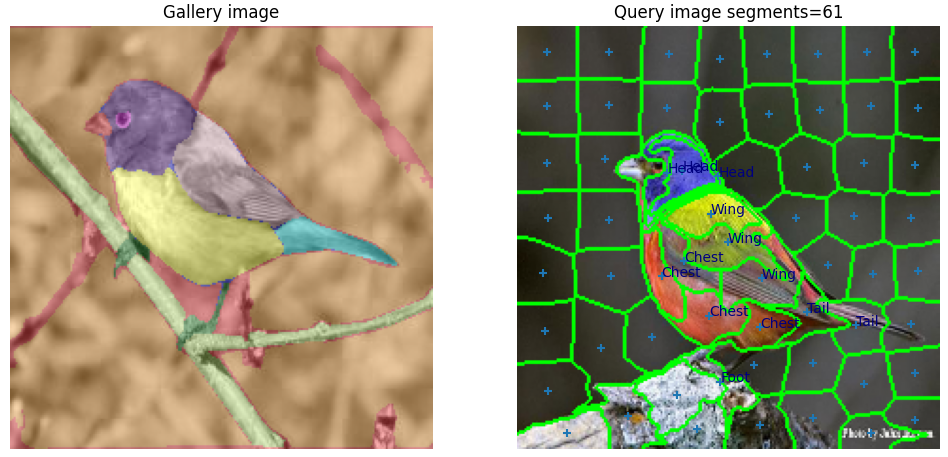}
   \caption{Transferred part labels to an unlabeled query image, after matching each superpixel gets a part label}
   \label{fig:transfer}
\end{figure}

%------------------------------------------------------------------

\begin{figure}[t]
  \centering
    \includegraphics[width=0.999\columnwidth]{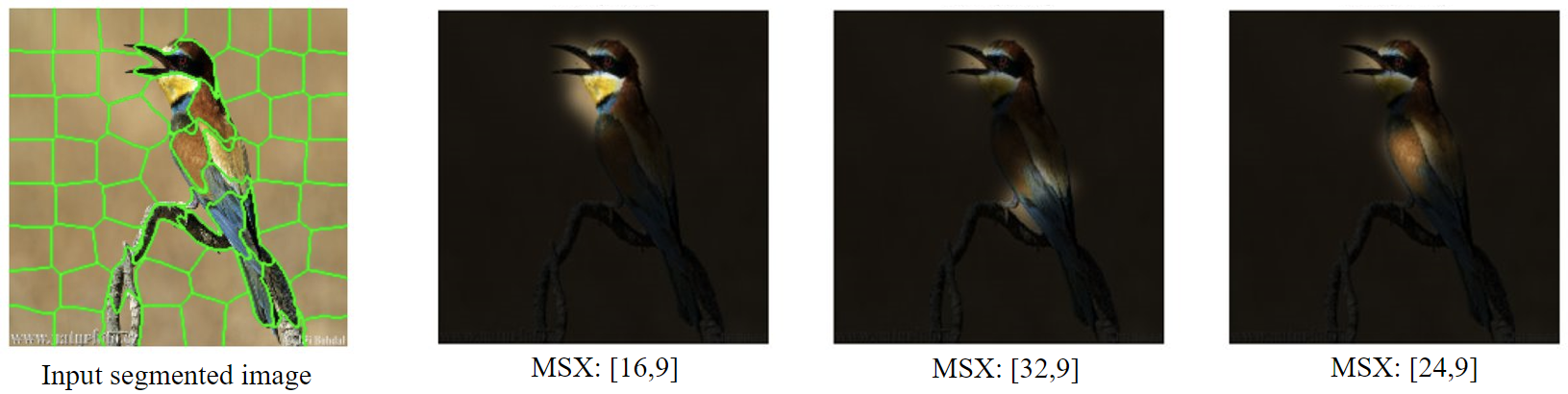}
   \caption{Three Minimal Sufficient Explanations for a single input image of Bee Eater class on the left for the Resnet50 model.}
   \label{fig:msxs}
\end{figure}

% -----------------------------------------------

\subsection{Computing Minimal Sufficient Explanations}
Our local explanations consist of a minimal set of regions that are sufficient for the model to produce a high score for the correct class. We call them Minimal Sufficient Explanations (MSXs).\footnote{We abbreviate Minimal Sufficient Explanation as MSX and not as MSE to deconflict it from mean squared error.} It has been shown that there are often multiple such MSXs for a single input image \cite{Shitole}. We extend the concept of MSXs to sets of superpixels found by SLIC and follow \cite{Shitole} to derive as many MSXs as we can find for each image using beam search.

Formally, the MSX of an image $x$ for the class $c$ w.r.t. classifier $\bm{\pi}$ is defined as a minimal set of superpixels that achieves a high prediction confidence $(\bm{\pi_c}(N_i) > P_h  \bm{\pi_c}(x))$ w.r.t. using the entire image, where we set $P_h=0.9$. That means that if the classifier is provided with a perturbed image where the rest of the image except the superpixels in the MSX are blurred out, it will produce a score of at least 90\% of the score on the original image. Furthermore, no strict subset of the superpixels in the MSX would give such a high score. 
For each image, the beam search identifies potentially multiple sets of superpixels as MSXs. Figure \ref{fig:msxs} illustrates the MSXs collected for a single input image with the Resnet50 image classification model.

\subsection{Computing global explanations}

After collecting MSXs for each image, we convert them to human-readable symbolic MSXs by mapping the superpixels in the MSX to the corresponding part labels, e.g., \textit{\{Bird-Head, Bird-Wing, Bird-Beak\}}. A symbolic MSX of an image identifies a minimal set of parts that is sufficient to recognize the class of that image by the given model and can be interpreted as a conjunction of propositions that represent part names. For a global explanation, we seek the shortest decision list of symbolic MSXs that collectively cover the data set.  

Since finding the shortest decision list is an NP-hard problem \cite{slavik1996tight}, we employ a greedy set cover algorithm \cite{young2008greedy}, which iteratively selects the symbolic MSX that covers most of the remaining images, until all image instances are covered.  Algorithm \ref{alg:algorithm} outputs a decision list which, although not guaranteed to be the smallest, can be interpreted as a propositional formula in monotone (positive literals only) disjunctive normal form, where each conjunct is a symbolic MSX in terms of part names. Collectively, the decision list covers all images in the input dataset. 

% ==============================================================
\begin{algorithm}[tb]
    \caption{Greedy Set Cover Algorithm}
    \label{alg:algorithm}
        \textbf{Input}: $\mathbb{U}$ - set of uncovered images; \\
        a family $S=\{S_1, S_2, \ldots, S_m\}$ of all symbolic MSX expressions. \\
        \textbf{Output}: $R$ - set of symbolic MSX that covers maximum images. \\
        \begin{algorithmic}[1]
        \STATE $R = \emptyset$.
        \WHILE{$\mathbb{U} \neq \emptyset$ and $i = 1, 2, \ldots, l$}
        \STATE select $R_i = \mathrm{argmax}_{j=1, \ldots, m} |S_j \cap \mathbb{U}|$
        \STATE $\mathbb{U} \leftarrow \mathbb{U} \setminus R_i$
        \STATE $R \leftarrow R \cup \{R_i\}$
        \ENDWHILE
        \RETURN $R$
    \end{algorithmic}
\end{algorithm}

\begin{table}
    \centering
    \begin{tabular}{lll}
        \hline{}
        Class\_id & Class Label & Label transfer accuracy   \\ 
        \hline
        n02690373 & airliner        & 0.9030   \\ 
        n01608432 & kite            & 0.9188   \\ 
        n04285008 & sports\_car     & 0.9504 \\
        n04612504 & yawl            & 0.8678  \\ 
        n03769881 & minibus         & 0.9091  \\ 
        n02089867 & walker\_hound   & 0.8410  \\ 
        \hline
    \end{tabular}
    \caption{Average accuracy of part label transfer for different classes in PartImageNet dataset}
    \label{tab:label_acc}
\end{table}

\section{Experimental Results}
\label{sec:experiment-description}

\subsection{Dataset description}

For producing part-based explanations, we utilized three different datasets. Stanford Cars dataset contains 16,185 images of 196 car classes \cite{krause2013collecting}, split into 8,144 training images and 8,041 testing images. Caltech-UCSD Birds 200-2011 (CUB-200) is an image dataset with 200 bird species and 11788 images \cite{wah2011caltech}. From each class in these two datasets, we created a condensed subset for annotation by randomly selecting 15\% of the images. Part annotations on these sample sets were performed on the Roboflow platform, where users could define labels and annotate each image as needed. For example, a bird image has been labeled with \textit{\{head, wings, tail, beak, eye, foot, chest, background\}} . We prioritize high-level features for improved correspondence accuracy among multiple meaningful concepts.
%Fortunately, we were able to 
We evaluated the part label transfer approach with existing part annotations in the PartImageNet dataset \cite{he2022partimagenet}, which contains 158 classes from ImageNet \cite{ILSVRC15} with approximately 24,000 images with part annotations per pixel.  We also used PartImageNet as gallery data to continue part transfer in the ImageNet dataset and generated explanations for different DNN models. The prelabeled PartImageNet data helped us evaluate the proposed method on a relatively larger dataset.

\subsection{Evaluation of label transfer}

We evaluated parts transfer using the label transfer accuracy metric \cite{liu2011nonparametric}. This metric measures the count of correctly predicted pixel labels against the total number of predicted pixels in an image. In our case,
we count the correctly predicted segment labels in lieu of the pixel labels. Table \ref{tab:label_acc} illustrates the performance of label transfer accuracy for a few categories using the HPF approach in the PartImageNet dataset. Overall, HPF using Resnet101 \cite{he2016deep} scores an average of 84.28\% on the part labels transfer task for 158 categories.

\begin{figure}[t]
  \centering
    \includegraphics[width=0.99\columnwidth]{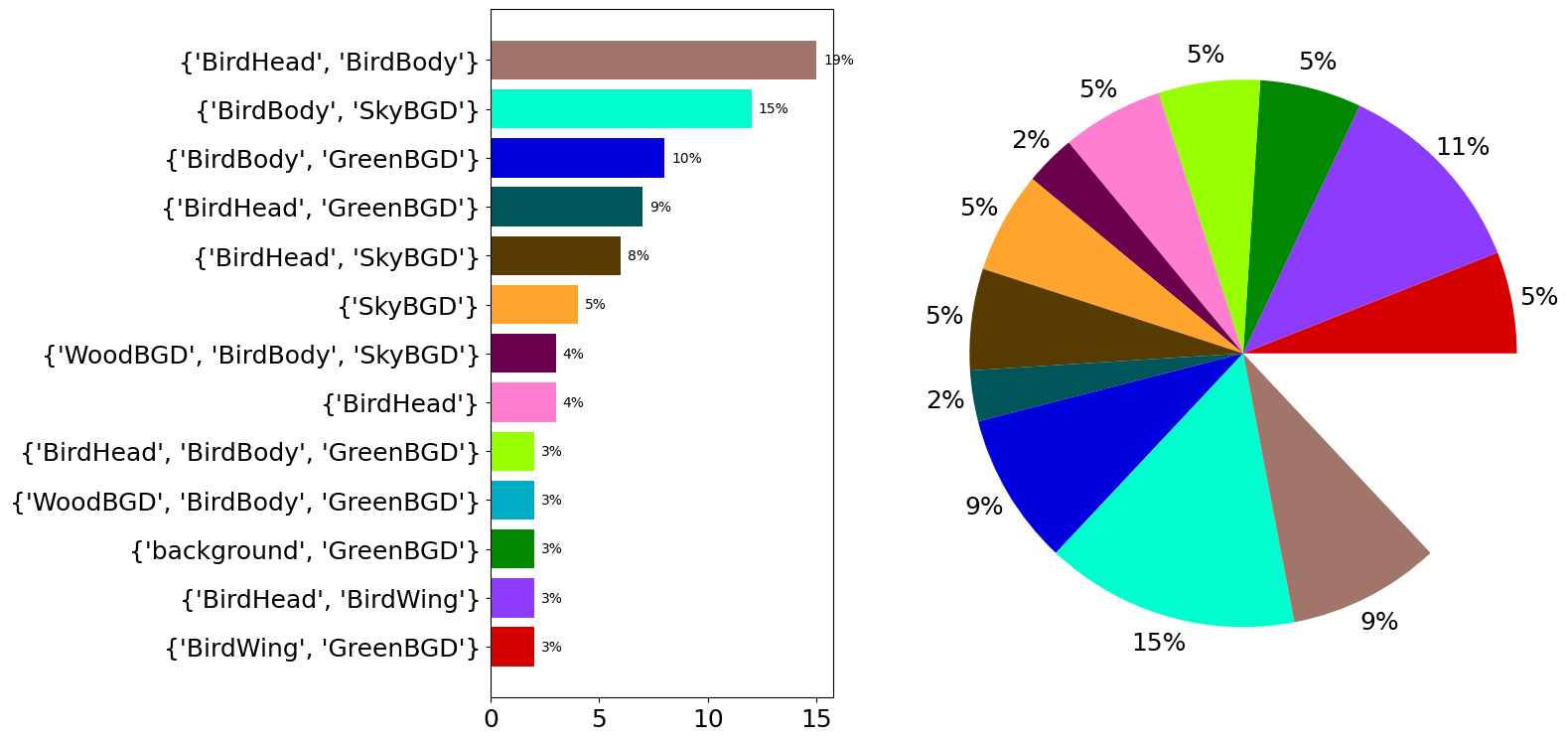}
   \caption{Global part-based explanations of Resnet50 classification model for the Bee Eater class followed by the proportion of test data that is explained by each rule.}
   \label{fig:setcover}
\end{figure}

%\vspace{-0.5in}

\begin{figure}
  \centering
    \includegraphics[width=0.99\columnwidth]{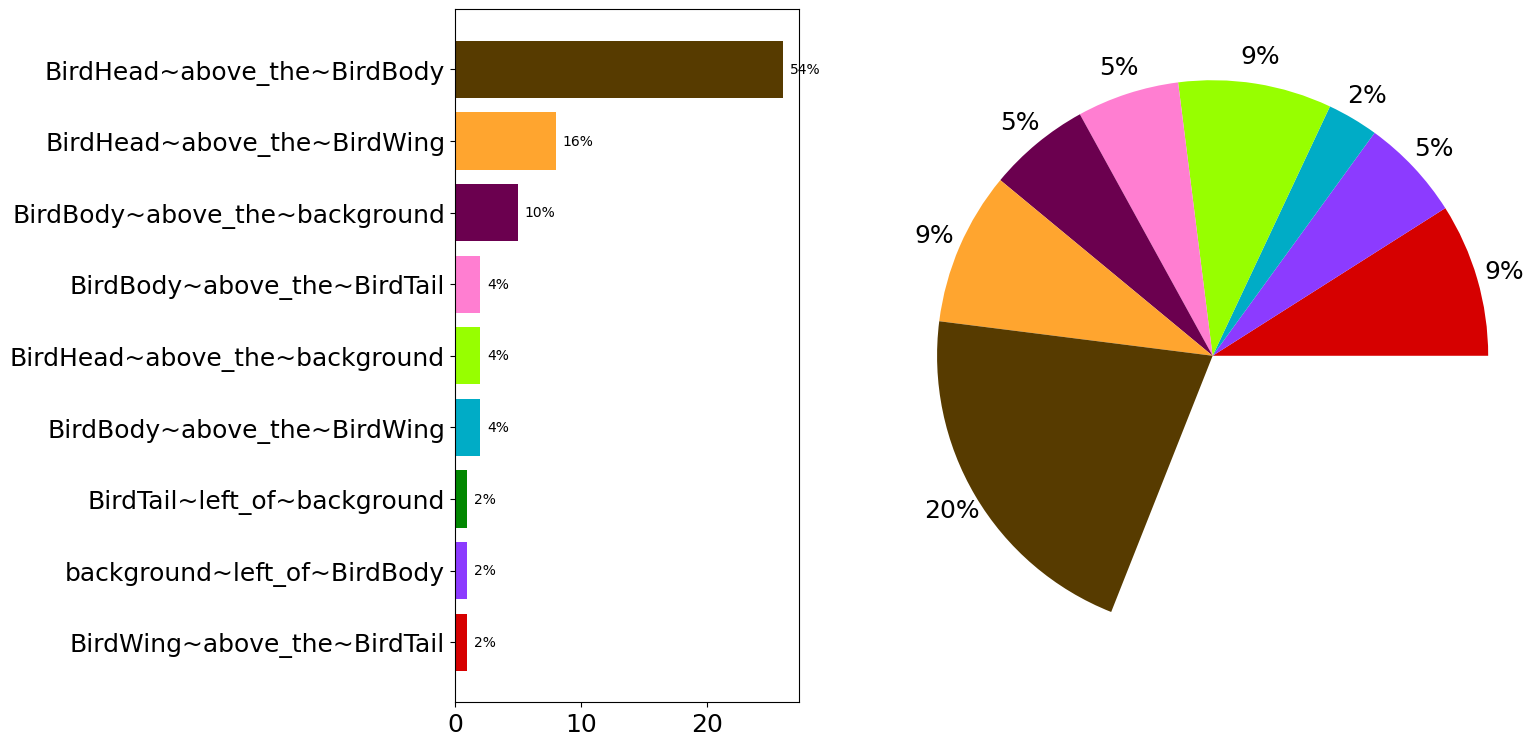}
    \caption{Global explanations of Resent50 model based on object parts and their relationships for the Bee Eater class followed by the proportion of test data that is explained by each rule.}
   \label{fig:setcover_spatial}
\end{figure}

% ============================================================
\begin{figure*}
\centering
% width=0.9\columnwidth]
\includegraphics[width=\textwidth]{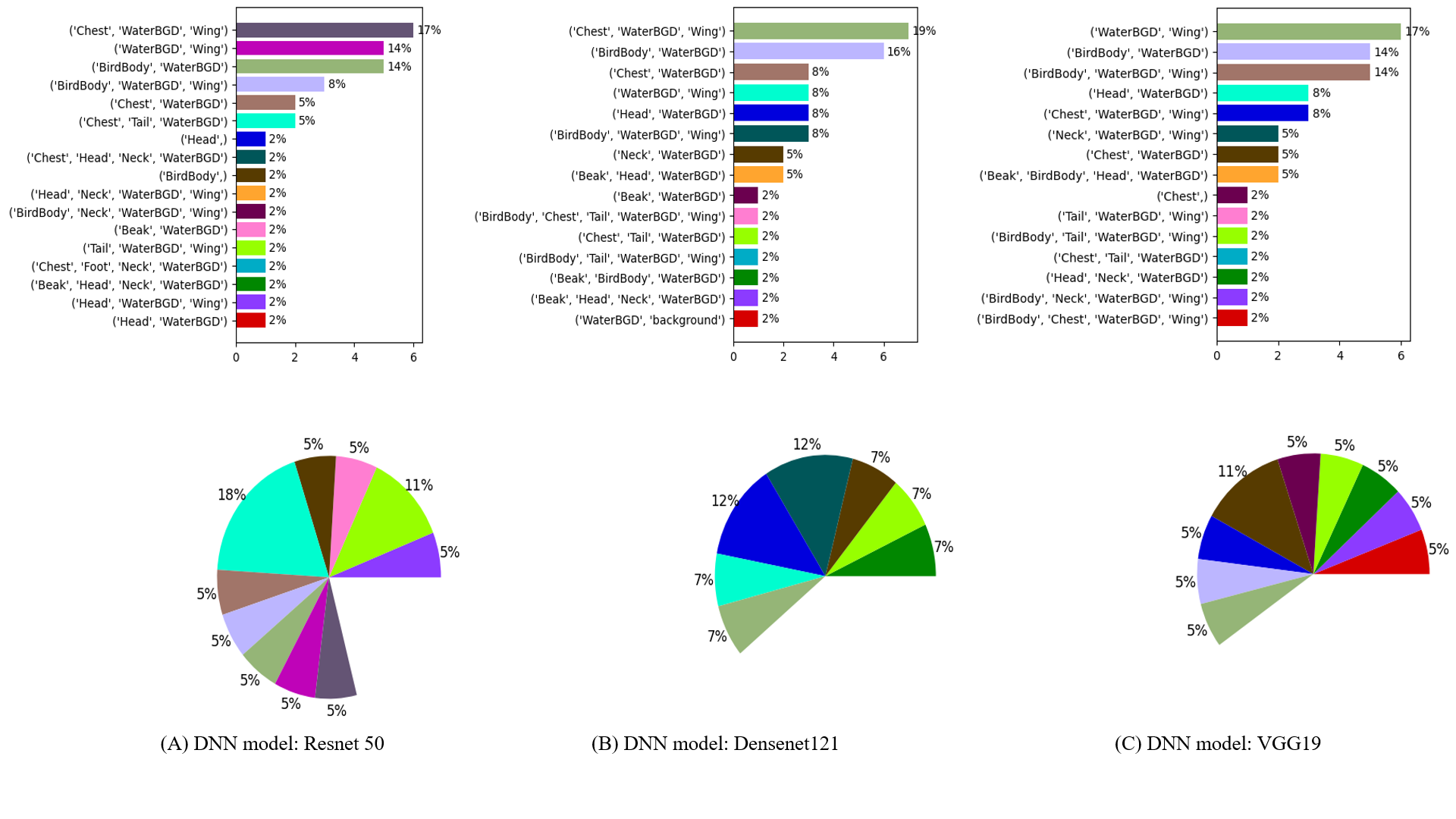} % 
\caption{Visualization of global explanations extracted over the same category data from different classifier models; these plots are generated over CUB200 dataset for the category of Black-footed Albatross.}
\label{fig:compare_models}
\end{figure*}
% ============================================================

\subsection{Evaluation of Global Explanations}
 
We split each dataset into 70:30 train-test sets within each class and generated global explanations as a decision list. The global explanations were evaluated for coverage over the test set, counting 
%, maintaining the order of the expressions in the decision list. 
each test example towards the first rule that covers it. 
Figure \ref{fig:setcover} shows the coverage of the rules in the global explanation, with the pie chart illustrating the proportions of test data covered by each symbolic MSX, color-coded to match the bar graph. Figure \ref{fig:compare_models} illustrates the global explanation in different model predictions, including Resnet50 \cite{he2016deep}, DenseNet \cite{huang2017densely}, VGG19 \cite{simonyan2014very} with symbolic MSX extracted from the train subset and the pie chart representing the proportions of test data covered by each rule.

In addition to rules based on part labels, we also experimented with rules that capture spatial relations between parts in the image. For example, a part-based rule like \{Bird-Head, Bird-Wing, Bird-Beak\} translates to relational rules such as \{(Bird-Head adjacent\_to Bird-Wing), (Bird-Head left\_of Bird-Beak)\}, etc. based on the relationship between the parts. 

We derived global explanations using greedy set cover over the relational rules. Importantly, relational rules may not have one-to-one correspondence with part-based rules. 

For instance, a part-based rule \{Bird-Head, Bird-Body\} might covers 19\% of images, but there might exist a relational rule \{Bird-Head above\_the Bird-body\} with 54\% coverage in the dataset, since it could be coming from multiple part-based rules like \{Bird-Head, Bird-Body\}, \{Bird-Head, Bird-Body, Bird-Tail\}, etc. 
Figure \ref{fig:setcover_spatial} illustrates the contribution of these relational rules to the model. 

\begin{figure*}
\centering
% width=0.9\columnwidth]
\includegraphics[width=\textwidth]{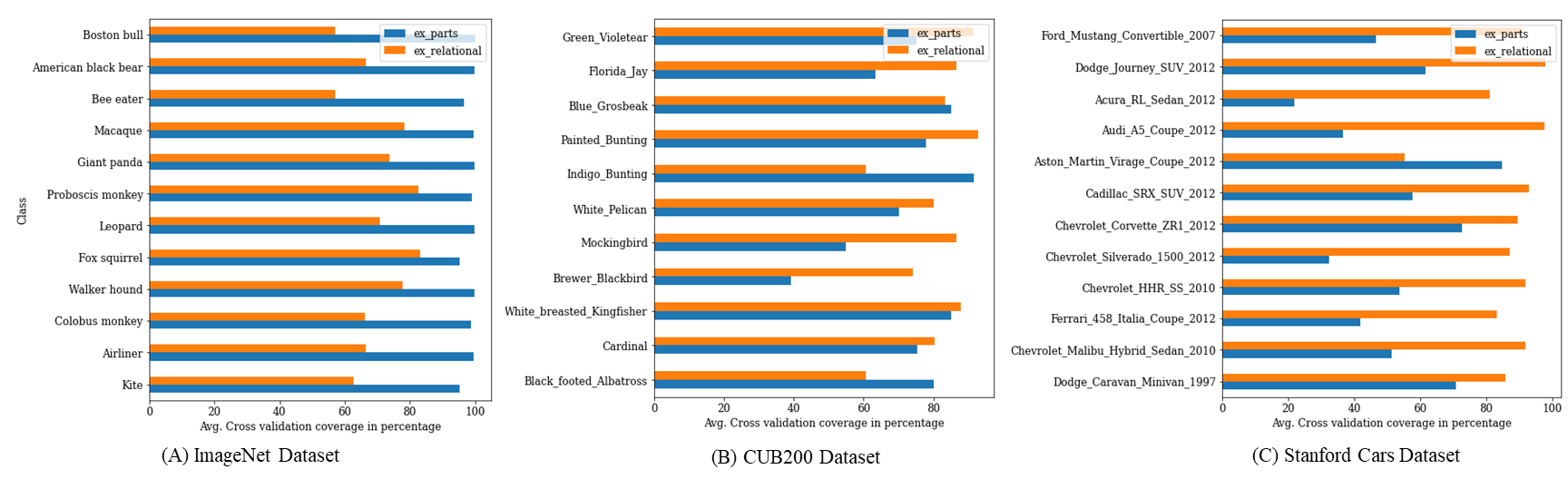} % 
\caption{Cross validation of coverage of Resnet50 predictions for a few classes in three different datasets: A) ImageNet, B) CUB200, C) Stanford Cars.}
\label{fig:cv_all}
\end{figure*}

We validated the global decision-list explanations using five-fold cross-validation w.r.t. different model predictions. Due to space constraints, we illustrate the validation of the explanation only from the predictions of the Resnet50 model. Figure \ref{fig:cv_all} displays the average percentage of images covered by the global part-based (\textit{ex-parts}), and relational explanations (\textit{ex-relational}). 
The relational explanations prominently uses ``above\_the" compared to the other spatial relational i.e. left\_of and adjacent\_to. We deliberately remove `background' relations from the explanations since it does not provide meaningful insights; however, this drops the coverage of relational decision lists compared to the part-based ones. Figure \ref{fig:cv_all} shows that relational expressions cover more images in CUB200 and Stanford datasets than the part-based ones, but not in ImageNet.

% ============================================================
% ============================================================
% ===================================================================

\section{Conclusions and Future Work}
\label{sec:results}

We presented an effective approach for obtaining global explanations of a Deep Neural Network model. These explanations are expressed in terms of human-interpretable part labels and their relationships. It is important to note that While global explanations shed light on what parts may be responsible for the model's decisions, they are not surrogate models in that the same explanation might hold for multiple classes.  

We envision future work that explores the extensions of our general approach of deriving global explanations from local MSXs to other tasks such as gene expression analysis, activity recognition in videos, and question answering from texts.

\section*{Acknowledgements}
This research was funded through the NSF grant CNS-1941892 and the ARO grant W911NF2210251.

\bibliographystyle{named}
\bibliography{ijcai24}

\end{document}